\documentclass[a4paper,11pt]{article}

\usepackage{fourier}
\usepackage{color}
 \usepackage{graphicx}
\usepackage{url}
\usepackage[affil-it]{authblk}
\usepackage{amsmath}
\usepackage{wrapfig}
\usepackage{xspace}

\usepackage[T1]{fontenc}
\usepackage{times}

\begin{document}

\title{Batch Normalization in the final layer of generative networks }

\author{Se\'an Mullery \&  Paul F. Whelan}
\affil{Vision Systems Group, School of Electronic Engineering, Dublin City University, Glasnevin, Dublin 9, Ireland}
\date{}
\maketitle
\thispagestyle{empty}

\begin{abstract}
Generative Networks have shown great promise in generating photo-realistic images. Despite this, the theory surrounding them is still an active research area. Much of the useful work with Generative networks rely on heuristics that tend to produce good results. One of these heuristics is the advice not to use Batch Normalization in the final layer of the generator network. Many of the state-of-the-art generative network architectures use this heuristic, but the reasons for doing so are inconsistent.
This paper will show that this is not necessarily a good heuristic and that Batch Normalization can be beneficial in the final layer of the generator network either by placing it before the final non-linear activation, usually a $tanh$ or replacing the final $tanh$ activation altogether with Batch Normalization and clipping.
We show that this can lead to the faster training of Generator networks by matching the generator to the mean and standard deviation of the target distribution's image colour values. 

\end{abstract}

\newcommand\blfootnote[1]{%
  \begingroup
  \renewcommand\thefootnote{}\footnote{#1}%
  \addtocounter{footnote}{-1}%
  \endgroup
}

\blfootnote{Article under IMVIP 2018 submission}

\section{Introduction}

Batch Normalization (BN) \cite{Ioffe} is considered one of the breakthrough enabling techniques in training deep neural networks. Without it, the gradient in each layer is tightly coupled to all other layers. Should the gradient on any layer be close to zero, then this chokes off the gradient to all subsequent layers during back-propagation of the gradient. This problem is known as vanishing gradients. BN works as follows. The activations from a layer, for a full mini-batch, are passed to the BN function. It calculates the sample mean and standard deviation for this mini-batch. It subtracts this mean and divides by this standard deviation to leave the activations for the mini-batch with a mean of zero and a standard deviation of one. Next, it does the reverse of this step by multiplying by a new standard deviation called $\gamma$ and adds a new mean called $\beta$. Importantly $(\gamma, \beta)$ are trainable parameters that exist for each channel of activations in a layer. The net effect is that the distribution of activations of one layer is shifted and expanded/contracted to match the input to the next layer. \\
Generative networks take various forms. \cite{Johnson2016} shows an example of image-in image-out for super-resolution and style transfer. Generative Adversarial Networks (GANs) introduced by\\ \cite{GoodfellowGAN2014} and Variational Autoencoders (VAE) use a latent vector $Z$ as input with the output being an image. 
All of these forms of generative networks must at some point constrain the activations to pixel values. For the case of colour images, the network must reduce down to three channels and the values would normally be constrained to be integers in the range $[0,255]$. To constrain the pixel values to be in an appropriate range the activation of choice is the $tanh$ although a $sigmoid$ could also be used. The $tanh$ function takes an unbounded real number and constrains it to the real number range $[-1,1]$. However as $tanh$ is a non-linear function and we see from Figure \ref{tanhFig} that inputs outside the range $[-2,2]$ will be saturated. When converted to an 8-bit image these saturated values will convert to colour values of $0$ and $255$. For most real images the pixel values will be well spread between $[0,255]$. If the generator network is to produce realistic looking images, it should naturally produce images that have pixel values well spread between $[-1,1]$ on the output of the $tanh$. The network previous to the $tanh$ should be aiming to produce activations with a mean close to zero and standard deviation close to one, to ensure a good spread of values entering the $tanh$. It is still reasonable for some activations of the $tanh$ to saturate. Many real-world colour values will be $0$ or $255$. 
Placing a BN layer between the final activations and the $tanh$ allows the activations earlier in the network to be less constrained. The BN will shift and spread/condense the values to a range that suits the $tanh$ function. Indeed we show that the $tanh$ may not be optimal and that BN with appropriate $(\gamma, \beta)$ values may suffice with simple clipping to $[-1,1]$, keeping in mind that clipping is a non-linear operation.



\section{State of the Art}
\subsection{Generative Adversarial Networks (GANs)}
GANs were originally introduced by 
\cite{GoodfellowGAN2014}. 
In their experiments on images, they are not explicit about the architecture design apart from to say the generator used $relu$ and $sigmoid$ activations while the discriminator used $maxout$ activations. There is no suggestion that BN was used. The ideas outlined in \cite{GoodfellowGAN2014} have sparked a large body of research, but the architectures used in most GANs for image generation follow the design of \cite{Radford2015}\\
\cite{Radford2015}, introduced what is commonly referred to as the DCGAN (Deep Convolutional GAN). They introduce the $tanh$ activation at the output of the final layer, observing that using a bounded activation like $tanh$ allows the model to saturate quicker and thus cover the colour space of the distribution. This is legitimate if the default output distribution is tightly compacted within the output activation, but if it is widely spread or far from centred then the $tanh$ may saturate the majority of the outputs making it very difficult for the generator to learn. \cite{Radford2015} advise using BN in most layers except for the final layer of the generator and the first layer of the discriminator. They noted that including BN in those layers led to sample oscillation and model instability which was avoided when it was removed. It should be noted here that we have also experienced this with the DCGAN using the original loss regime from \cite{GoodfellowGAN2014}. However, with some other designs and loss regimes, we find that BN is a benefit and that this heuristic may not be appropriate everywhere.\\ 
\cite{Goodfellow2016} refers to the key insights of the DCGAN, stating that BN is left out of these layers so that the Model can learn the correct mean and scale of the distribution. BN has learnable parameters ($\beta$, $\gamma$) that can represent these and are condensed into ($\beta$, $\gamma$) though there may be reasons the \cite{GoodfellowGAN2014} loss regime prefers to distribute this over the rest of the weights in the network.  A clear explanation of why BN in these specific layers causes oscillation and instability has to our knowledge not been resolved. One of the key problems with GANs that follow \cite{GoodfellowGAN2014} is that the loss functions do not inform us how training is progressing.\\
\cite{ArjovskyWGAN2017} introduced a new loss function to GANs, called the Wasserstein distance. The Wasserstein distance conveys how training is progressing, though in this first implementation it was crudely approximated in a computerised setting by means of constraining the weights of the discriminator network. In their experiments, they remove all BN from DCGANs in their entirety. They also used a constant number of filters in each layer instead of the doubling at each layer used in \cite{Radford2015}. There doesn't seem to be any reasoning for this design choice, except perhaps to show that the Wasserstein distance can overcome all these handicaps or perhaps that they do not matter. 
\cite{Gulrajani} improves upon the work of \cite{ArjovskyWGAN2017} with the improved WGAN. The approximation in a computerised setting was the main improvement though this also led to necessary changes in architecture. \cite{Gulrajani} do use BN in the generator but use Layer Normalization \cite{LayerNorm2016arXiv160706450L} in the Critic (WGAN name for the discriminator). It is not stated whether they use BN in the final layer of the generator but given that they say they follow the DCGAN architecture of \cite{Radford2015} we can assume that they do not. The GitHub repository they supply certainly does not appear to use it in the final layer. The \cite{Gulrajani} architecture will be referred to as the iWGAN in this paper.
\subsection{Image-in Image-out networks}
\cite{Johnson2016} created a generative network for super-resolution and style-transfer. They used a pre-trained VGG network \cite{simonyan2014very} as their loss function, and the generative network consisted of multiple residual convolutional blocks with BN. However they also intentionally leave BN out of the final layer stating that they use a $tanh$ function on the output layer to ensure that the pixels are in the range $[0,255]$. The suggestion here is that the $tanh$ replaces the need for BN, though there is no explanation as to why one would be a direct replacement for the other. In section \ref{BNSection} we will show the different effects of $tanh$ and BN and give the case that BN makes more sense in the output of a generator network.

\subsection{How to evaluate generative models}\label{EvalGen}
It is worth considering how to evaluate the performance of any type of generative model. \cite{Theis2015} compare average log-likelihood, Parzen window estimates \cite{Breuleux2009} and visual fidelity of samples which at the time were the most commonly used evaluation criteria for generative models. They show robust theoretical reasons why these are largely independent of each other. Their results show that Parzen window estimates should in general be ruled out. Of average log-likelihood and visual fidelity they show that both can be misleading. Plausible samples can be generated while still achieving poor log-likelihood. However good log-likelihood can be achieved with no guarantee of visually plausible samples.\\

We can see from many of the above state of the art in generative networks that the heuristic to leave BN out of the final layer of generator network is widely used. Despite this the justifications given for this appear inconsistent. In the following sections we will show for some of the solutions above that BN in the final layer can lead to faster training. 

\section{How BN relates to $tanh$ }{\label{BNSection}}
\begin{wrapfigure}{r}{0.4\textwidth}
  \vspace{-20pt}
  \begin{center}
    \includegraphics[width=0.35\textwidth, height=0.28\textwidth]{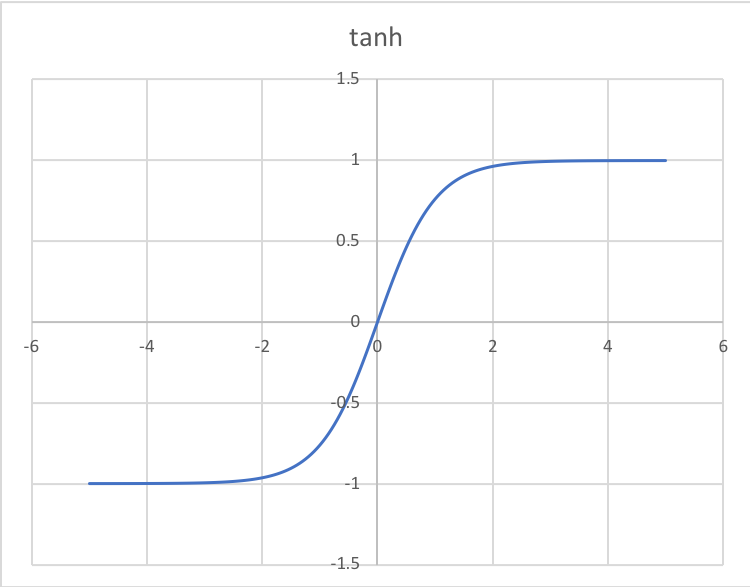}
\end{center}
\vspace{-20pt}
  \caption{The $tanh$ function.}
  \label{tanhFig}
  \vspace{-0pt}
\end{wrapfigure}
Consider what happens to a distribution of activations that vary beyond the bounds $[-2,2]$ in Figure \ref{tanhFig}. Any activations beyond these bounds will saturate to $[-1,1]$. While the output will be bound to the desired $[-1,1]$ the network has no way to differentiate between any one saturated activation and another, which creates difficulties with learning. We have seen in practice that this can be overcome, and the values do eventually begin to fall in the correct range, but this can take time. 
Now consider what happens if we put BN before this $tanh$. In the first part of the BN algorithm, the mean and standard deviation of the batch of activations will be calculated, the mean will be subtracted from each activation value, and then each will be divided by the batch's standard deviation. The expected value of any activation is now zero, which is the central value of the input to the $tanh$ function and the standard deviation of an activation is one which means that most activation values will be between $[-1,1]$. The majority of activations (up to two standard deviations) will result in an unsaturated output from the $tanh$ functions. From two standard deviations and above we expect saturation. This assumes that BN's learnable parameters do nothing, but as they are trainable, they can vary to suit the target distribution. If any channel of the target distribution has a non-zero mean, then the $\beta$ parameter for that channel can vary to suit. Likewise for the standard deviation of colour values of the target distribution. They must change in such a way that they move the activations to a value that will give the desired target distribution after output from the $tanh$. This does present a problem as $(\gamma, \beta)$ are parameters of an affine operation and varying them cannot undo the effect of a non-linear function like $tanh$.\\
This brings up the question, why the $tanh$ at all? If the BN can learn or be set to the target distribution for each channel is there any benefit to the $tanh$? We show later that $tanh$ is unnecessary and it can, in fact, bring a small improvement to replace it with a clipping operation. \\
It should be noted that BN is not a replacement for a non-linearity in general. BN is an affine operation so it cannot serve as a non-linearity. Instead, we are saying that the affine operation it offers along with a clipping activation (itself a non-linearity) is of more utility here than the $tanh$.
Aside from the effects of adding complexity and generalisation to the network a $tanh$ or clipping non-linearity is necessary here or it can cause artefacts in the generated image where a number outside the range $[0,255]$ is inserted into the image. 

\section{Method}
As outlined in section \ref{EvalGen}, evaluation of generative models is not straightforward. In this paper, we will only be looking at the first phase of training. To some extent, we will use visual fidelity but usually only in conjunction with a histogram of the image showing the distribution of values. As \cite{Theis2015} work predates the work of \cite{ArjovskyWGAN2017, Gulrajani, Johnson2016} they did not consider using the loss functions of these methods as a way to measure the performance of generative models. We do use these loss functions, but it should be noted that they suffer the same drawbacks that average log-likelihood suffer. Namely, that plausible samples can be generated while still achieving high loss and a low loss can be achieved with no guarantee of visually plausible samples.

\subsection{Experiments} 

To determine the efficacy of using BN in the final layer of the generative network, we will look at the  Image-in Image-out architecture of \cite{Johnson2016} and the iWGAN architecture of \cite{Gulrajani}.
We use the Keras framework with the TensorFlow back-end.\\
For the Image-in Image-out architecture we consider super-resolution. We will compare the histogram/images created at the initial phase of training and also look at the trend of the perceptual loss function, which is a meaningful value that should tend to zero as training progresses. We use a subset of ImageNet \cite{imagenet_cvpr09} (19439 images) of size $288 x 288$ as our high-res target distribution. From this set, we make a low-res set of images of size $72 x72$. The Generator network must learn to output the hi-res data set in response to the low-res set input. To see how training is progressing we will take an image from outside the target data set and resize to $72 x 72$. This will be passed to the generator at periodic intervals of training. The output hi-res image will be considered in terms of visual fidelity and histogram distribution. 
Using this underlying architecture, we compare the combinations of $tanh$ alone, BN with $tanh$ and BN with clipping. For BN in conjunction with the $tanh$ we use the default initialisation values of $(\gamma = 1, \beta = 0)$ as these are reasonable numbers for spread across the $tanh$ function. For BN-with-clipping we initialise the $(\gamma, \beta)$ with the per-channel standard deviation and mean of the hi-res training set.\\
For the iWGAN architecture, we use the critic loss, which \cite{ArjovskyWGAN2017} showed was proportional to the Wasserstein distance, as a measure of how quickly our different output layers converge. CIFAR-10 \cite{CIFAR10} $32 x 32$ pixel images of frogs (5000 images) are used for training.  
Once again we will consider visual fidelity and in particular the histogram spread of values 
As above, we compare the combinations of $tanh$-alone, BN-with-$tanh$ and BN-with-clipping. 
Details of both architectures and hyper-parameter values are in the Appendix.


\section{Results}

\begin{figure}
 \begin{center}
    \includegraphics[width=0.8\textwidth]{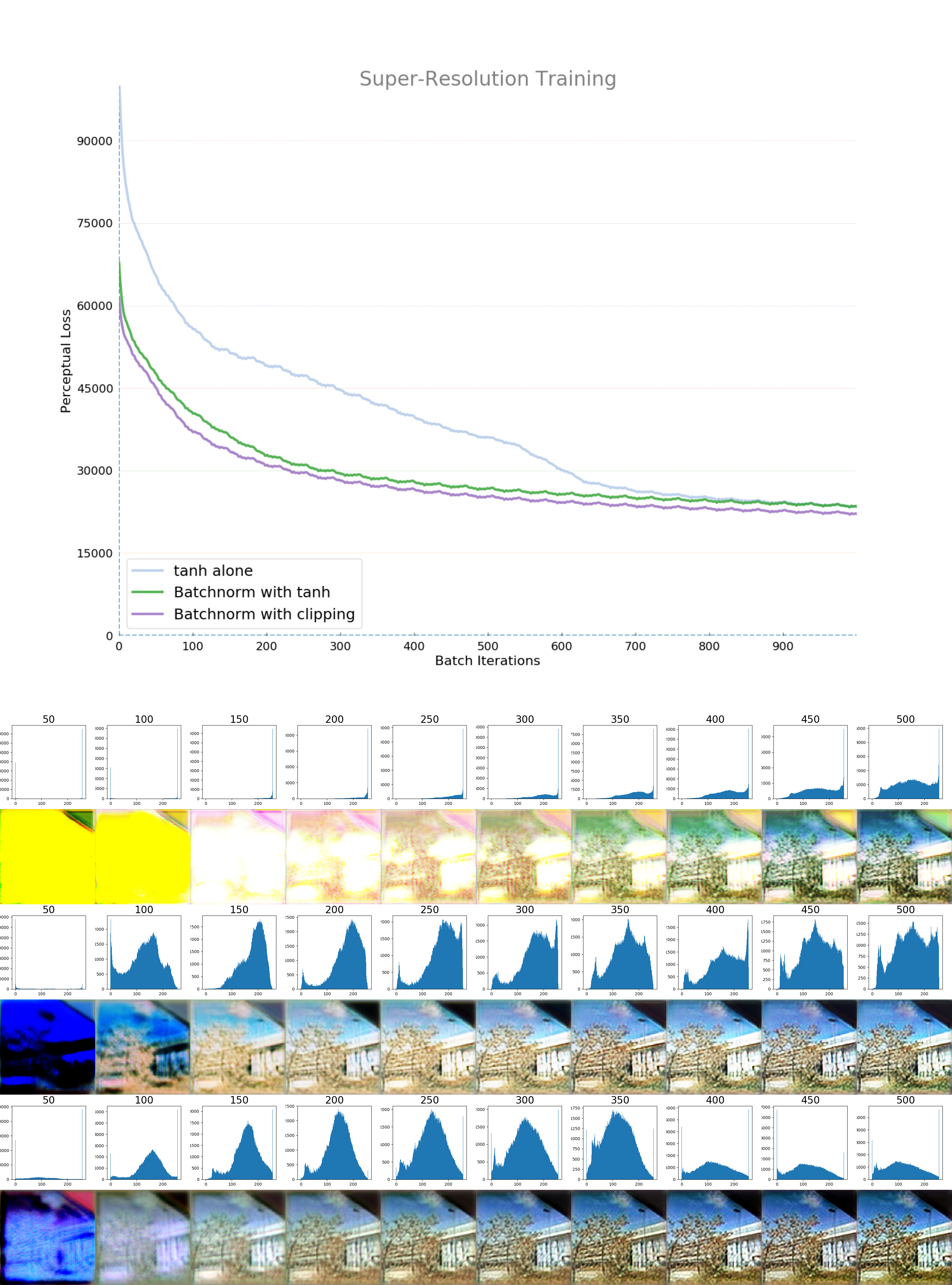} 
    \caption[]{
    Super Resolution: Note: the black triangle in the top right-hand corner is the overhang of a building and is not an artefact. This image was used to determine if the outputs had any issue dealing with a large section of saturated (to black in this case) pixels. The first row of histograms relates to the $tanh$-alone. Note, that above the histograms are shown in the batch update iteration at which they were recorded. The second row portrays BN-with-$tanh$ and the third-row show BN-with-clipping. For these, we show the histograms and image examples for generator iterations 100-1000. The perceptual loss is a large magnitude number as it scales with the size of the activation layers used in the loss network; VGG in this case.\\ Code and results can be found at \url{https://github.com/seanmullery/Super-Resolution}
    }
    \label{SRTraining}
  \end{center}
\end{figure}

\begin{figure}
 \begin{center}
    \includegraphics[width=0.8\textwidth]{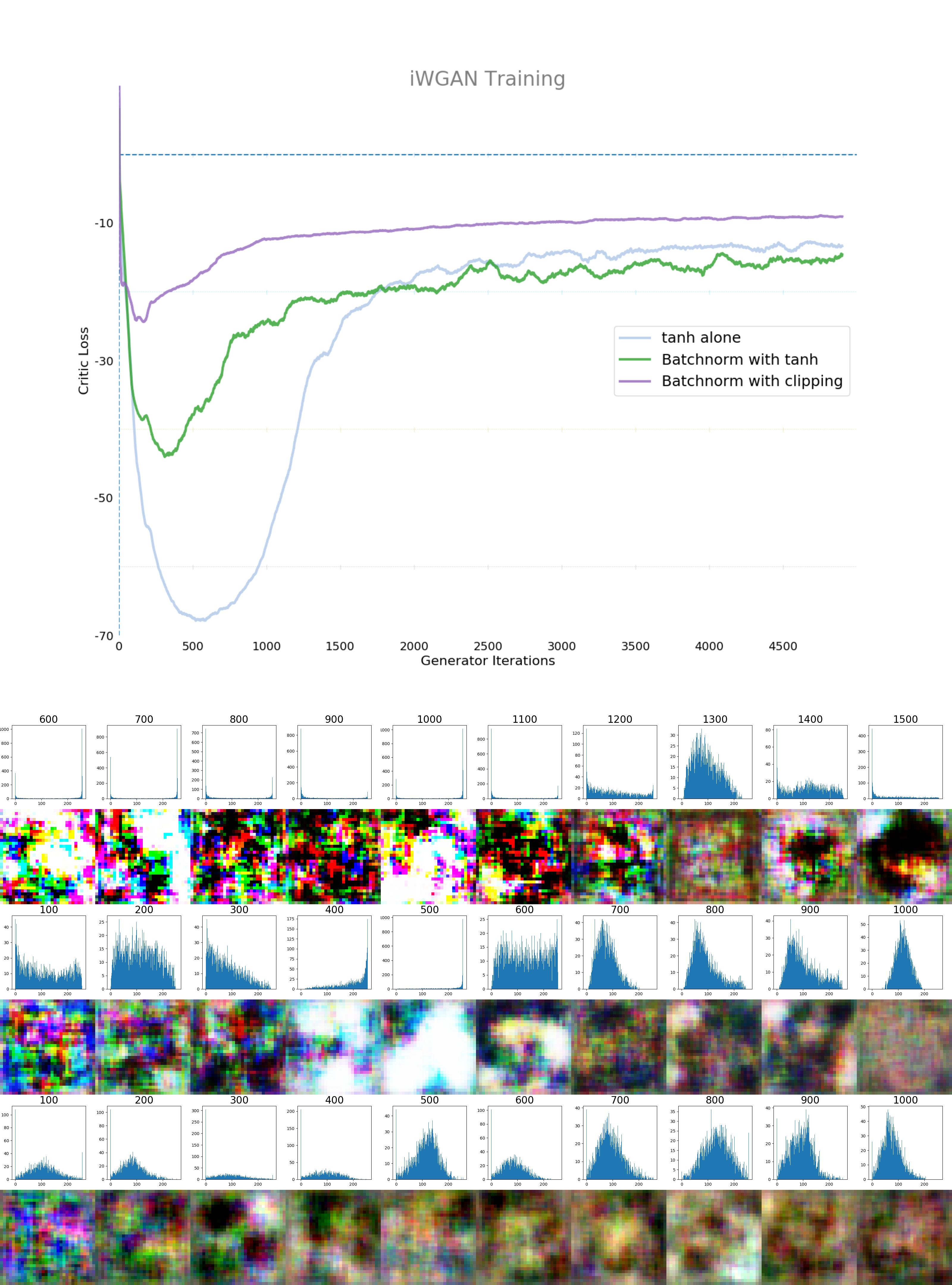} 
    \caption[]{iWGAN: The Wasserstein distance is normally shown in this way (as negative). The closer to zero the smaller the distance and the closer to convergence. The first row of histograms relates to the $tanh$-alone. Note, that above the histograms are shown the generator iteration at which they were recorded. For the $tanh$-alone, we show from 600-1500 to show that reasonable histograms only start to appear from 1200 onwards. The second row portrays BN-with-$tanh$ and the third-row show BN-with-clipping. For these, we show the histograms and image examples for generator iterations 100-1000. The images are meant to be of frogs but it is very early in training, so this would not be obvious at this stage.\\ Code and results can be found at {\url{https://github.com/seanmullery/iWGAN}} }
    \label{iWGANTraining}
  \end{center}
\end{figure}

Figure (\ref{SRTraining}) shows three plots of perceptual loss at each batch update iteration for the three final layer types, $tanh$-alone, BN-with-$tanh$ and BN-with-clipping used with the super-resolution network. \\
Figure (\ref{iWGANTraining}) shows the critic loss (Wasserstein distance) change over generator iterations. The three plots show the distance for $tanh$-alone, BN-with-$tanh$ and BN-with-clipping used with the iWGAN design. All plots use a 100 point running mean, the actual loss has a much greater variance. 

\section{Discussion}
Figure (\ref{SRTraining}) shows the perceptual loss function for the three scenarios, $tanh$ alone, BN with $tanh$ and BN with clipping. We can see that BN-with-clipping shows a very slight improvement on BN-with-$tanh$. It is difficult to determine if choosing a better initialisation for the BN parameters $(\gamma, \beta)$ (from the default) for the BN with $tanh$ would close the gap or whether this is a more fundamental difference between using BN with clipping and $tanh$. We can certainly see that in the early part of training the architectures with BN outperform $tanh$ alone. As training progresses, $tanh$ eventually converges with the others. This is something we see with the iWGAN as well. However, BN has predominantly used as a method of faster training. Here as well we achieve faster an smoother convergence in the early part of training while the $tanh$ alone version must learn the correct mean and standard deviation of the distribution. We can see the effect of this on the visual fidelity of the output images. The first two rows show the $tanh$ alone, the middle two the BN with $tanh$ and the final two rows show BN with clipping. A large part of the loss relates to the pixel values not being in the correct dynamic range. We also see that the network is still able to cope with an image that has a large section of saturated pixels.\\
Looking at Figure (\ref{iWGANTraining}) we see all curves have a steep change in critic loss but at different times. For the $tanh$-alone the steep reduction in loss occurs between 900-1500 generator iterations. For the network with the extra BN layer before the $tanh$ the reduction begins much earlier at approx 500 iterations and the Wasserstein distance never reaches the magnitudes shown in the example without the BN. A look at the image output and more importantly the histograms for the $tanh$ alone (top two rows) shows that the steep reduction coincides with the histogram falling in line with the expected histograms of the data set. Note the number of generator iterations at the top of each histogram to show where it was recorded. This suggests the vast majority of the Wasserstein distance relates to the generated image having most pixels in saturation.
For the BN with $tanh$ (middle two rows) this also happens earlier suggesting that the large change in Wasserstein distance is caused by moving the image pixel values to match the dynamic range that allows them to produce a well-behaved histogram. For the BN with clipping, where the $(\gamma, \beta)$ values are already set to the standard deviation and mean of the target distribution we see only a minimal change and we note the histograms are well behaved from very early in training. The images should be of frogs. This is not at all obvious, and yet we see that most of the Wasserstein distance does not relate to perceptual content but instead to the spread of colour values.

\section{Conclusion}
We have shown that the use of BN in the final layer of generator networks deserves reconsideration. While the heuristic to remove BN from the final layer may be justified when using the architectures and techniques of \cite{GoodfellowGAN2014, Radford2015}, its use in image-in image out generation and Wasserstein GANs may be beneficial, at least regarding the early part of training. Consideration should also be given to replacing $tanh$ on the output with BN with clipping particularly if the mean and standard deviation of the target data set can be reliably approximated. \\
The question of whether BN in the final layer can lead to a better long-term result is still unclear, and further long-run testing would be required for this. This may require better methods for assessing the quality of generative networks. The issues surrounding the problems encountered by \cite{Radford2015} deserve further investigation. If BN in the final layer is the reason for the oscillation and instability, it would be interesting to learn why. There may be other ways to solve the problem that give better results than simple removal of the BN from the final layer.

\section*{Acknowledgments}
The principal author would like to acknowledge Jeremy Howard and Rachel Thomas for the fast.ai MOOC, and Shaofan Lai on which some of the base programs for the experiments in this paper were based.

\appendix

\section{Appendix: Network and Hyper-paramter details}

For Super-Resolution: VGG networks are used as the loss networks with a weighted average $[0.1,0.8,0.1]$ of the VGG layers [Conv1\_1, Conv2\_1, Conv3\_1] respectively. We use ADAM \cite{DBLP:journals/corr/KingmaB14} with Keras default learning rates. The generator is as that used by \cite{Johnson2016} in their supplementary material, table 2, 4x super-resolution\\ \url{https://cs.stanford.edu/people/jcjohns/papers/fast-style/fast-style-supp.pdf}\\
For iWGAN:The training regime is that the critic runs 5 iterations for every generator iteration apart from in the first 25 generator iterations where the critic runs 50 iterations for every generator iteration. In order to further ensure that the critic remains at optimality, the critic runs 50 iterations on the 500th iteration of the generator. The latent vector $Z$ has dimension 64. 
We use ADAM with learning rate 0.0001, $\beta_1=0.9$ and $\beta_2=0.99$. 
The network architecture can be seen at {\url{https://github.com/seanmullery/iWGAN}}

\bibliographystyle{apalike}

\bibliography{references.bib}

\end{document}